\documentclass[runningheads]{llncs}
\usepackage{graphicx}
\usepackage{amsmath}
\usepackage{amsfonts}
\usepackage{booktabs}

\usepackage{framed,color,xcolor}
\usepackage{multirow}
\usepackage{nicefrac}
\usepackage{threeparttable}
\usepackage{pgfplots}
\usepackage{soul}
\usepackage{microtype}
\usepackage{graphics}
\usepackage{framed,color,xcolor}
\usepackage{multirow}
\usepackage{nicefrac}
\usepackage{threeparttable}
\usepackage{todonotes}
\usepackage{multirow}
\usepackage{colortbl}
\newcommand{\customer}{user}
\newcommand{\Customer}{user}
\usepackage[T1]{fontenc}
\usepackage[utf8]{inputenc}
\usepackage{babel}
\usepackage[font=small,labelfont=bf]{caption}

\begin{document}
%
%\title{Contribution Title\thanks{Supported by organization x.}}
\title{Machine Translation Customization via Automatic Training Data Selection from the Web}

\titlerunning{Machine Translation Customization via Training Data Selection}
%\author{Thuy Vu\orcidID{0000-0003-1056-6975} and Alessandro Moschitti\orcidID{0000-0003-2216-8034}}
\author{Thuy Vu and Alessandro Moschitti}
\institute{Amazon Alexa AI,\\ Manhattan Beach, California, USA \\
\email{\{thuyvu,amosch\}@amazon.com}
}
\maketitle              % typeset the header of the contribution
\begin{abstract}
Machine translation (MT) systems, especially when designed for an industrial setting, are trained with general parallel data derived from the Web. Thus, their style is typically driven by word/structure distribution coming from the average of many domains. In contrast, MT customers want translations to be specialized to their domain, for which they are typically able to provide text samples.
We describe an approach for customizing MT systems on specific domains by selecting data similar to the target customer data to train neural translation models.
We build document classifiers using monolingual target data, e.g., provided by the customers to select parallel training data from Web crawled data. Finally, we train MT models on our automatically selected data, obtaining a system specialized to the target domain.
We tested our approach on the benchmark from WMT-18 Translation Task for News domains enabling comparisons with state-of-the-art MT systems. 
The results show that our models outperform the top systems while using less data and smaller models.

\end{abstract}
\section{Introduction}

Industrial MT services have greatly impacted multiple commercial applications, e.g., Google Translate and Amazon Translate.
It has also become an indispensable technological component worldwide during the current pandemic to disseminate COVID-19's public service announcements to the public~\cite{Coronavirus2020}.
The result has been collectively attained by leveraging Web data: training examples (parallel text) can indeed be automatically built by aligning sentences from multilingual pages, which naturally occur on the web ~\cite{buck-koehn:2016:WMT2,dirtcheap,uszkoreit-EtAl:2010:PAPERS,cda}.

The harvesting of parallel data from the web has been shown successfully by~\cite{banon-etal-2020-paracrawl,dirtcheap}, resulting in highly heterogeneous collected data, as sampled from the entire web. Thus, the distribution of the content is inevitably dominated by the commercial websites working in a multi-language setting.
On the one hand, this distribution may reflect the average expected demand submitted to an MT service by web {\customer}s; 
on the other hand, it can hardly capture the specificity of less represented domains.
In particular, {\customer}s working with domains that traditionally do not require multilingual content, e.g., documentation of local administration or businesses having no internationalization interest, may find a general-purpose translation inadequate.

For example, if we use general terms, such as \emph{project meeting} and \emph{sport meeting}, which occur in many websites, a standard MT system provides rather accurate Italian translations, \emph{incontro di progetto} and \emph{incontro sportivo}, respectively.
However, if we try terms less frequent in multilingual web data, for example, \emph{condo meeting} or \emph{condominium meeting}, we may obtain the following wrong translations: \emph{riunione del condominio} or \emph{condominio incontro}, instead of the correct one, \emph{riunione di condominio} \footnote{As of May 2020, Google Translate provided \emph{riunione condominiale}, which, although correct, is  a bit too formal term for this kind of meeting.}.
In particular, the MT system cannot select the right preposition \emph{di} since
    (i) the most typical Italian construction uses \emph{del}, and
    (ii) \emph{condo meeting} is infrequent in web parallel data. 
In contrast, \emph{project meeting} is correctly translated in \emph{incontro di progetto} by most MT services: we did not observe mistakes of the type \emph{incontro del progetto} or a less used term \emph{incontro progettuale}. We speculate that such term, being more frequent, is typically supported by more training examples.

Current MT systems deal with the problem of under-represented domains by averaging the patterns observed in all available domains.
Thus, the bias in generating translation towards the populated domain persists.
This causes a translation targeting low-frequent phrases to use irrelevant or inappropriate words.
In extreme cases, such problems may create embarrassing biased translations~\cite{PoliticoBias2018}; for example, \emph{pornographic domains} appear very frequently on the web~\cite{7536508}, if not adequately filtered, common terms may be interpreted in a sex key.

This paper explores automatic customization/personalization of MT systems by automatically selecting training data \emph{similar} to the text in a target customer application.
Such data will carry terminology and syntactic constructions specific to the target domain.

Our main assumption, supported by general machine learning theory, is that we can customize neural network models by training them with this selected data. % similar to the one of the target domain.
Such an approach can produce three main benefits:

\begin{itemize}
\item The MT system requires less data to learn to translate in the target domain than when using general data. Indeed, specific domains are characterized by less lexical variability due to the need to express specific concepts/situations. The use of less data produces efficiency benefits at training time, with possibly a better translation quality in the domain.

\item The fine-tuning step with customized domain data can increase accuracy in translating text from such domain in neural MT. In particular, infrequent patterns with respect to the average web distribution will better emerge from the model in the target domain as they will occur relatively more often.

\item A positive side effect of this approach is that specific data can automatically diminish the bias on undesired domains, e.g., political inclinations or explicit content, when operating in a critical setting, e.g., kid protected content. Indeed, amplifying the term distribution of the kid domain can help mitigate the impact of very different and undesired training data.

\end{itemize}

To customize an MT system on a target domain, we assume to know the monolingual data of the domain in advance.
This is a realistic assumption as the customer can specify their target data/domain, e.g., providing their website or textual documentation.
Simultaneously, the MT service provider can continue to refresh their parallel data repository asynchronously and periodically.
Therefore, the \emph{customization} process is reduced to selecting the parallel data portion similar to the one from the target domain to train/fine-tune the MT models on the target context. We propose the design of topical classifiers to recognize the target domain data among the extremely large web crawled data.
We note three important aspects:
\begin{itemize}
\item First, the data provided for the customization domain does not need to be parallel.
We only need monolingual text data similar to the target domain to train the topic classifier.
This is very important, as acquiring parallel data can be a key limitation to any customization approach's applicability. In contrast, monolingual data can be easily acquired from the customer's website, documentation or other related data.

\item Our classifier is built to predict webpages instead of sentences as carried out in previous MT domain adaptation works based on language model~\cite{D11-1033}.
Using entire pages allows for reaching a high accuracy in selecting data potentially similar to the target data since the document content distribution is not sparse and richer than the content of individual sentences.

\item The negative examples can be generated by randomly sampling webpages from the entire crawled data. Indeed, given the very low occurrence probability of the documents of the target domain in comparison with billions of pages in the crawled data, the number of false negatives would be extremely low.
\end{itemize}

We tested the following research questions:
\begin{enumerate}
\item[$\mathbf{q}_1$]: Can we build efficient document classifiers to select large training data for MT systems specific to target domains?
\item[$\mathbf{q}_2$]: Are the classifiers  accurate enough to select training data for the target domain from web crawled data?
\item[$\mathbf{q}_3$]: Does the data selected by the classifiers produce improvement of the MT systems when tested on the target domain?
\end{enumerate}

To answer the questions above, we compared our selection approach against the state-of-the-art MT systems of the WMT-18 News Translation benchmark. The results show that using the data selected by our classifier, we can train a much simpler model and still be on par with the state-of-the-art approaches, e.g., those proposed by RWTH and Microsoft Research. These use a Big Transformer and are much more expensive.
Our results show that (i) our approach for selecting target data is effective; and (ii) it is possible to customize MT systems on a target domain, i.e., the news domain. Although  wider experimentation over different domains of possibly different sizes is needed to claim that our is a general-purpose approach to MT customization and personalization, our paper provides examples in such directions, enabling promising future work. 
It also shows interesting evidence on the potential of IR techniques for converting web data in specific applications without going through knowledge-based methods.

\section{Domain Customization Approach}
Our approach consists in
  (i) acquiring monolingual data for a target domain;
 (ii) training a topic classifier for such domain, using the acquired data as positive examples and randomly sampled web data as negative examples;
(iii) selecting parallel data of the target domain by applying the built classifier to the monolingual text part of the crawled data;
 (iv) training or fine-tuning the MT system on the data selected by the classifier; and finally
  (v) applying the trained MT system for user data.

We describe the details in the following subsections.

\subsection{Components and Notation}

Our model requires the following components:

\begin{itemize}
\item a general large repository $\mathcal{C}$ of crawled parallel data for MT training.
\item Several domains $D^+_1,..,D^+_n$ for different applications, businesses, and users.
\item A sampling procedure $S$ to get the negative examples from $\mathcal{C}$ not in $D^+_i$, denoted $D^-_i=S\left(\mathcal{C},D^+_i\right)$.
\item A linear fast topic classifier $R_{D_i}$, which we will train on $D_i=\{D^+_i, D^-_i\}$.
\item A vanilla state-of-the-art MT model, $T_{\mathcal{C}}$, to be trained on parallel data.
\end{itemize}

The customized MT system will then be $T_{C_i}$, trained on $C_i \subset \mathcal{C}$, where $C_i = R_{D_i}(\mathcal{C})$.
Specifically, $R_{D_i}$ selects relevant parallel data from $\mathcal{C}$ based on ${D_i}$ characteristics.
Note that $R_{D_i}$ is trained using $D^+_i$ as positive examples and $D^-_i=S(\mathcal{C},D^+_i) \subset \mathcal{C}$ as negative examples.

\subsection{Customization Pipeline}
Figure~\ref{diagram} describes our pipeline to build an MT system customized for a particular user/domain.
The diagram displays three different processes: (i) the training of a classifier $R_{D_i}$, (ii) the data selection, (iii) the MT training, and (iv) the customized translation.

In the first phase, the {\Customer} provides a sample of the \emph{Target Data} constituted by monolingual documents.
These are positive examples (blue squares) used to train a classifier for the target data.
The negative examples (grey squares) are sampled from the \emph{Heterogeneous Dataset} (parallel data crawled from the web).

  \begin{minipage}[b]{0.40\textwidth}
   \hspace{-2.5em}
	\includegraphics[width=1.3\linewidth]{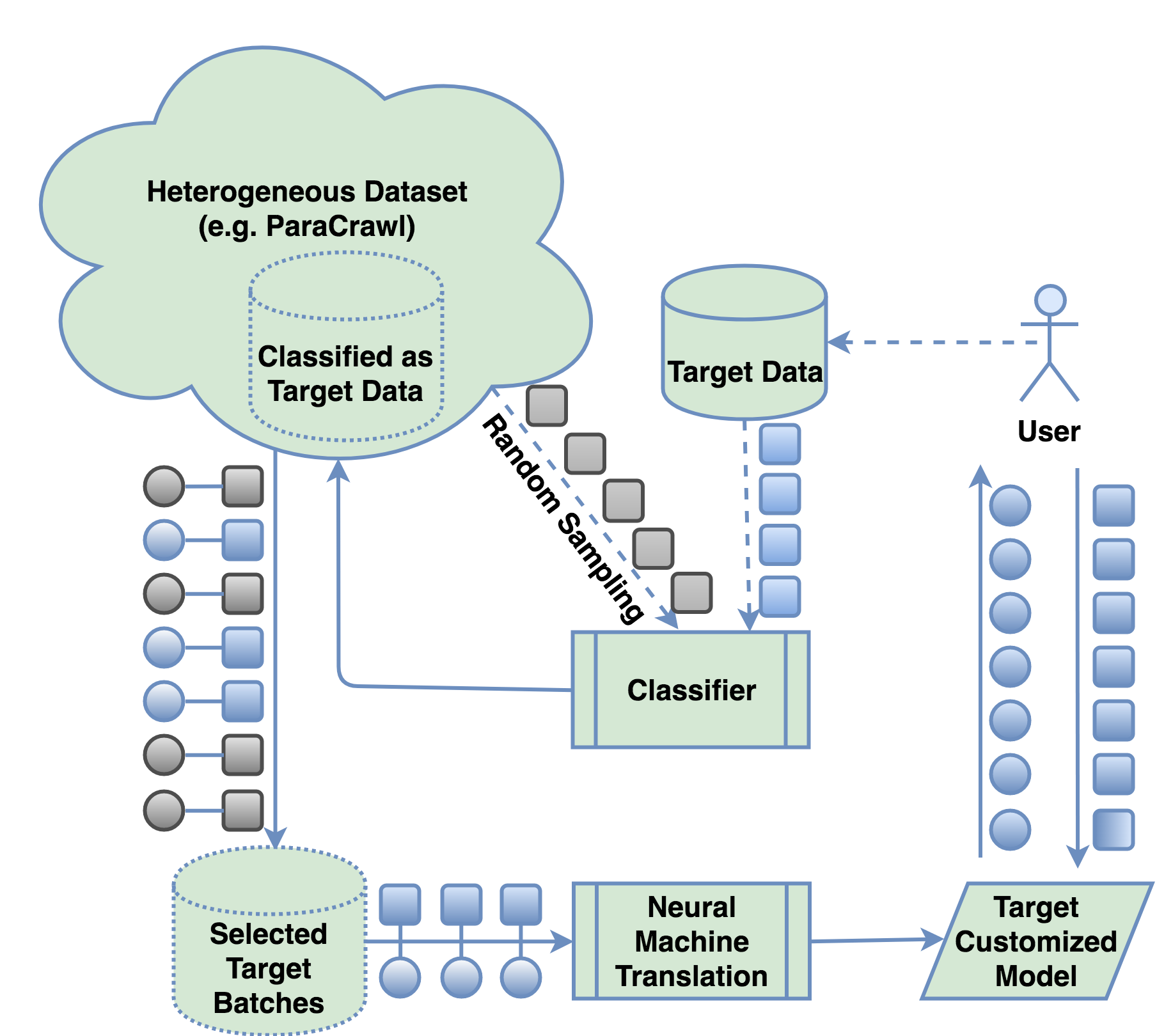}
	\captionof{figure}{\normalsize Customization process of MT Systems}
	\label{diagram}
    \vspace{1em}
  \end{minipage}
\hspace{2em}
  \begin{minipage}[b]{0.45\textwidth}
    \begin{tabular}{l c}
 	\toprule
 	Corpus & Sent. (MM) \\
 	\midrule
	 News Commentary v13 & 0.3 \\
 	Rapid (press releases) & 1.3 \\
 	Common Crawl & 1.9 \\
 	Europarl v7 & 2.4 \\
 	\hline
 	ParaCrawl (Zipporah) & 40.6 \\
 	ParaCrawl (BiCleaner) & 27.7 \\
 	\bottomrule
    \end{tabular}
    \vspace{.3em}
    \captionof{table}{\normalsize Training data for WMT-18 for English--German}\label{wmt-18-data}
    \vspace{1em}
    \end{minipage}

In the second phase, the trained classifier produces a classification score for all \emph{Heterogeneous Dataset} documents.
The classification is done by exploiting only the monolingual side of the parallel data (in the same language of the target domain data).
Although the Heterogeneous Dataset  can be potentially very large, the classifier runs in linear time and can be parallelized.

In the third phase, the pairs of parallel documents, i.e., the circle and square pairs, are ranked with respect to the classifier score.
The top $k$ \emph{Selected Target Batches} are split in pair of parallel sentences, and used to train the Neural MT model.
Note that using ranked data we (i) avoid to tune up a classification threshold, which can be rather challenging as it requires the annotation of crawled data; and (ii) can select higher quality data from the top until we need or until the MT system does not improve anymore.

Finally, the users can apply the \emph{Target Customized Model} (MT system) on their new monolingual text and receive translated data.

\vspace{-1em}
\subsection{Target Data Classifier}
\vspace{-.5em}

As we need to process millions of instances, we implement our standard text classifier with Support Vector Machines (SVMs).
As previously mentioned, the positive examples are created by randomly sampling a fixed amount of text from the target data provided by the customer.
In contrast, the negative examples are randomly sampled from the heterogeneous background dataset.

The instance representation is based on the bag-of-word model, using the weighting scheme for the terms described below.
Given a document $d$, the term frequency $tf$ of a word $\omega_i \in d$ is normalized by the following equation:
$$tf \left(\omega^n_1,d\right) = \frac{{count}\left(\omega^n_1,d\right)}{\max_{(\overline{\omega}^n_1, \overline{d}) }{count}\left(\overline{\omega}^n_1, \overline{d}\right)}$$
where, $count\left( \omega_i, d \right)$ is the number of $\omega_i$ occurrences in $d$.

In general, the classifier scores indicate the likelihood of a text sampled from a source to be in the same domain of the target data.

\vspace{-.5em}
\subsection{Selection Approach}
\vspace{-.5em}
\label{ranker}
In principle, a binary topic classifier would be appropriate to select relevant data. However, estimating the threshold associated with an effective F1 could be cumbersome as we do not have a development set reflecting the target data required by the MT system. Thus, we do not even know the amount of the needed data and the Precision required to  train the MT system effectively.
Therefore, instead of a classifier, we use a ranker. 
This can be formally defined as a function $$R:\mathcal{C} \rightarrow \mathcal{P}(\mathcal{C}),$$ which takes the set of documents, $\mathcal{C}=\{d_1,..,d_{|\mathcal{C}|}\}$, and returns a subset of size $k$, i.e., $R(\mathcal{C}) = [d_{i1},...,d_{ik}]$.
To implement the reranker, we can still use a binary SVM classifier, which will learn a point-wise reranker: this outputs a score $s(\vec d)=\vec w \cdot \vec d + b $. 
The ranker is supposed to compute the set of indices as $[{i1},...,{ik}]=\text{k-argmax}_i\hspace{.3em} s(\vec d_i)$, where $\text{k-argmax}$ returns the indices of the top scored $k$ documents.

$R$ selects domain data from a heterogeneous dataset (e.g., the crawled data) based on the classifier's scores when applied to the monolingual documents.
The top $k$ documents associated with their parallel counterparts are selected for training, or fine-tuning, the MT systems.

\vspace{-1em}
\section{Experiments}
\vspace{-.5em}

We demonstrate the effectiveness of our proposed method step-wise in a typical pipeline to build state-of-the-art MT models using data selected by our proposed classifier.
For this purpose, we first study the performance of the domain classifier separately.
We then show its concrete impact in training both standard MT systems and a large-scale well-known MT benchmark, the WMT-18 News Translation Shared Task.
This experiment enables us to explain empirically the performance of our approach in comparison with other MT systems trained on the exact benchmark setting and using the same experimental dataset.
The setting includes a large, noisy parallel data crawled from the web.

\vspace{-.5em}
\subsection{Experimental Setup}
\vspace{-.5em}
We use the evaluation setting of the News Shared Task from WMT-2018~\cite{bojar-EtAl:2018:WMT1}.
In particular, we carry out experiments on two translation tasks: English--German and German--English.
\vspace{-.5em}

\subsubsection{Data}
The data provided by WMT-2018 is summarized in Table~\ref{wmt-18-data}.
The first four datasets are considered of high quality or \emph{clean} in this experiment.
The next two datasets, newly introduced as part of the WMT-2018 benchmarks, are ParaCrawl cleaned by two different filtering methods.
They are parallel sentences extracted automatically from crawled web data and subsequently cleaned by Zipporah and BiCleaner.
     
In our experiment, we propose the following setting to implement our diagram in Figure~\ref{diagram}:\vspace{-.5em}
     \begin{itemize}
        \item The \verb+News Commentary v13+'s text in English side is used as Target Data as we set news translation as the target domain application.
        \item The \verb+ParaCrawl (BiCleaner)+ data is considered as the Heterogeneous Dataset, given its web nature, large size, and noise quality.
        \item Our neural MT models are trained with all clean data (the first four datasets) in Table~\ref{wmt-18-data} and an automatically selected portion from the Heterogeneous Dataset.
    \end{itemize}
\vspace{-.5em}
 
It should be noted that this data comes in the form of individual paired-sentences. We simulated documents by grouping sentences in batches to train our document classifier.
The procedure is a key factor as we can (i) avoid possible topical bias regarding individual documents but (ii) also capture sufficient thematic or stylistic information of the target domain. In other words, we do not classify individual sentences but sentence batches.

\vspace{-1em}
\subsubsection{Domain Classifier Data}
\vspace{-.5em}

\label{classifier}

We generate positive and negative examples for building a classifier for news domains as follows:
\vspace{-.5em}

\begin{itemize}

    \item for positive examples, we form an example by randomly selecting $n$ English sentences without repetition from the news data, \texttt{News Commentary v13}.
        The example may contain sentences from different source documents yet they are from the news domain.
        This helps capture the journalistic signal in news reports while discouraging possible topical text from a particular story or section.

    \item For negative examples, we alternatively sample from the \texttt{ParaCrawl} dataset cleaned by \texttt{BiCleaner} while keeping the size of $n$ sentences per example.
        Even though journalistic text can appear in the example, the probability with respect to all the other content of the web makes the contribution of false-negative examples negligible.
        
    \item We also set the ratio between negative/positive to 2:1 to have enough positive examples.

\end{itemize}

\vspace{-.5em}
\subsection{Domain Classifier Results}
\vspace{-.5em}

We study the performance of the proposed classifier in this section.
Specifically, we set $n$ to 100 for the number of sentences per example.
This results in 2,828 and 5,656 positive and negative examples, respectively, from \texttt{News Commentary v13}.

We apply a split of 30\% for training and 70\% for testing.
As the original sentences from \texttt{News Commentary v13} are distinct, the generated examples for training and testing should also share no content overlapping.
We used SVMs to build the classifier/reranker.
We set the probability parameter to enable Platt scaling calibration on the classifier score.
The feature set consists of 70,000 most frequent words with stop-words being removed in the dataset.

  \begin{minipage}[b]{0.50\textwidth}
   \hspace{-2em}
      \resizebox{1\linewidth}{!}{
\begin{tikzpicture}
	\tikzset{mark options={mark size=.8, line width=.8pt}}
	\begin{axis}[
	        	name=plotDev,
	        	scale=1.0,
		xlabel={\Large $n$ sentences per batch},
		ylabel={\Large classifier accuracy},
		grid=both,
		xtick={0, 10, 20, 30, 40, 50},
		xmax=55,
		ytick={0.6, 0.7, 0.8, 0.9, 1.0},
		ymin=0.58,
		ymax=1.02,
		ylabel near ticks,
		xlabel near ticks,
		ylabel style={yshift=-1ex},
		style={/pgf/number format/.cd,fixed,precision=3},
		tick label style={font=\large},
		xlabel style={yshift=1ex},
		width=10cm,height=5cm
		]
	\addplot [color=blue, mark=x]
	    table[x=size, y=accuracy,] {data/batch-size-accuracy.txt};
	\end{axis}
\end{tikzpicture}
	}
%	\vspace{-.3em}
	
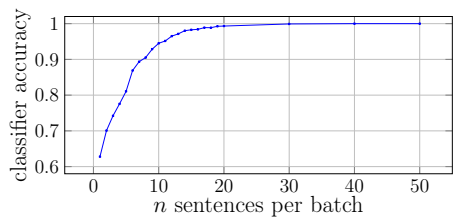
\captionof{figure}{\normalsize Accuracy of the classifier \\ in different setting of $n$.}
	\label{fig:different_n}
  \end{minipage}
  \hfill
  \hspace{-1.7em}
  \begin{minipage}[b]{0.5\textwidth}
    \centering
    \resizebox{1\linewidth}{!}{
    \begin{tabular}{rc}
    \toprule
                                     & Accuracy \\ \hline
 Sentence-based Classifier           & 62.8\%   \\% \hline
 Batch-based Sentence Majority       & 77.8\%   \\ \hline
 Batch-based Classifier (Our Method) & 99.0\%   \\% \hline
    \bottomrule
    \end{tabular}
    }
    \vspace{1em}
    \captionof{table}{Accuracy comparison of the \\ proposed method and other baselines.}\label{classifier_performance}
   \vspace{.5em}
    \end{minipage}
   \vspace{1em}

We use the default setting for the other  SVM parameters of the \verb+sklearn.svm+ toolkit. We compare the effectiveness of our proposed selection method, \emph{Batch-based Classifier}, with two related yet different configurations as baselines:
\begin{itemize}
    \item \emph{Sentence-based Classifier}: we build a classifier similar to the above configuration, except for the size $n$ of each batch set to $1$.
    This is equivalent to building a classifier, where the documents are constituted by just individual sentences.
    \item \emph{Batch-based Sentence Majority}: we classify a batch of $n=100$ sentences via majority voting, i.e., we apply the \emph{Sentence-based Classifier} to all sentences of the batch, and we classify the batch according to the majority of positive or negative classifications.
\end{itemize}

The accuracy of the classifier and the baselines is presented in Table~\ref{classifier_performance}.
Training and classification at document level is much more advantageous than the one at sentence level.
Because the word distribution from a larger text is more statistically reliable -- the basic theory of large samples provides support for such intuition, where \emph{the samples} in our case are constituted by set of words.
Note that the distribution of positive and negative batches is still 1:2, i.e., the same sentence distribution; thus the results are comparable.

To better show the intuition that the larger is the sentence batch, the higher is the accuracy, we have plotted the accuracy of our batch classifier with respect to the batch size in Figure~\ref{fig:different_n}.
We see that as soon as the batch content is larger than 10 sentences, the accuracy exceeds 95\%.
With batches of 20 sentences or more, the classifier reaches perfect accuracy.
This can be explained by the fact that random documents from the Web (approximated by the ParaCrawl) are statistically very different from those of the target domain.
At the same time, we built our training and test sets with a positive/negative example distribution of 1:2.
The classification accuracy over the entire ParaCrawl, which shows a much more skewed distribution can be significantly lower.
However, the purpose of this experiment was to show that we can build an accurate classifier.
Given the above positive result, we can use the classifier for reranking our data.
The effectiveness of the classifier in selecting data will be shown in the next sections.

\hspace{-1em}
  \begin{minipage}[b]{0.41\textwidth}
\resizebox{1\linewidth}{!}{
\begin{tabular}{ r r r r r }
\toprule
\multicolumn{1}{ c }{ParaCrawl} & \multicolumn{2}{ c }{Buckets} & \multicolumn{2}{ l }{Clean \& Bucket} \\ %& \multicolumn{2}{c}{Clean \& Cumul.} \\ %\hline
  & 2017 & 2018 & 2017 & 2018 \\ %& 2017 & 2018 \\
\midrule
 0\%  & -- & -- & 27.2 & 32.4 \\ %& 27.2 & 32.4 \\% \hline
{\bf 0\%--25\%}  & {\bf 28.1} & {\bf 34.3} & {\bf 29.8} & {\bf 36.2} \\ %& 29.8 & 36.2 \\% \hline
25\%--50\%  & 23.4 & 27.8 & 27.3 & 32.8 \\ %& 28.8 & 34.9 \\% \hline
50\%--75\%  & 12.7 & 14.7 & 25.2 & 30.3 \\ %& 28.3 & 34.4 \\% \hline
75\%--100\% & 5.8 & 6.6 & 25.0 & 29.7 \\ %& 28.2 & 34.4 \\% \hline
\midrule
0\%--100\% & 23.7 & 29.21 & 28.22 & 34.41 \\ %& 28.2 & 34.4 \\% \hline
\bottomrule
\end{tabular}
}
\captionof{table}{BLEU-based Evaluation of CSE on WMT-18}\label{wmt-18-result}
\end{minipage}
%\hspace{0em}
  \hfill
  \begin{minipage}[b]{0.51\textwidth}
%    \centering
\hspace{-0.8em}
\resizebox{1.05\linewidth}{!}{
\begin{tabular}{l r r r r}
\toprule
System                  		&  EN-DE &  DE-EN \\
\hline
RWTH Aachen             	&    --    	&  0.413 \\
Microsoft Research      	&  0.551 	&    --    \\
University of Cambridge 	&  0.537 	&  0.395 \\
University of Edinburgh 	&  0.352 	&  0.261 \\
JHU MT Systems          	&  0.377 	&  0.317 \\
Universitat Polit{\`e}cnica de Val{\`e}ncia &  -- &  0.321 \\
\hline
ONLINE-A                &  0.561 &  0.346 \\
ONLINE-B                &  0.396 &  0.310 \\
ONLINE-C                &  0.060 &  0.268 \\
ONLINE-D                & -0.385 & -0.296 \\
ONLINE-E                & -0.416 & -0.074 \\
\bottomrule
\end{tabular}
}
    \captionof{table}{Average-$z$ of Human Evaluation Scores for WMT-18 Systems, Including 5 Anonymized Translation Services.}\label{humanw18}

    \end{minipage}

\subsection{Machine Translation Results}

We study the impact of the proposed data selection approach in MT tasks.
In particular, we conducted experiments to address the following two questions:
\begin{itemize}
\item    [(i)] Can the classifier select relevant data for the target domain?
\item    [(ii)] Can the selected data be used to improve the state-of-the-art  in MT on a specific domain?
\end{itemize}

To reliably answer the second question, we used the WMT-18 benchmark as it is well-known both in academic and industrial MT communities. 
We performed two main experiments: the first aims at exploring the quality of the candidates with respect to their position in the rank generated by the topic classifier.
The second aims at measuring the potential of our selected data with respect to the state of the art.

\subsubsection{Data Quality in the ranked examples}

In these experiments, we used an efficient MT approach, namely, the LSTM cell by \cite{D15-1166,bahdanau+al-2014-nmt}, as we were interested in relative values of the accuracy and carrying out a fast experimentation.

We order documents and thus sentences in ParaCrawl in the descendent order of the classifier score described in Sec.~\ref{ranker}.
We then split the rank into four buckets of the same size.
We used one bucket at a time to train an MT model using the default setting of Sockeye~\footnote{\url{https://github.com/awslabs/sockeye}~\cite{DBLP:journals/corr/abs-1712-05690}} (LSTM cell).
We evaluated such models against the standard WMT-2017 and WMT-2018 test sets, using BLEU as our evaluation metric.
The results are reported in Table~\ref{wmt-18-result}, under the column \emph{Buckets}, using the evaluation tool, \verb+sacrebleu+~\cite{W18-6319}.
Each row, labeled with an interval percentage, corresponds to a different MT system trained with the rank interval data. 
As expected, the systems trained with higher ranked data show a larger BLEU score.
The system trained with the bottom bucket shows a very low performance.
It is also interesting to compare with the second column showing the results using the 6M clean sentence pairs from WMT-2018: the MT system trained with our selected data in the first interval, 0\%--25\%, shows a higher accuracy. This is important as the crawled data is generally rather noisy, meaning that our classifier can select clean MT data.

Additionally, we combined the bucket data with the clean WMT-2017/2018 data.
The results are reported under column \emph{Clean \& Bucket}, starting from the second row.
We note that the combination can improve the system using just the clean data, e.g., from 29.8  to 36.2 on the WMT-2018 test set.
This confirms that our approach can improve MT systems.
The combination of clean data with all the other buckets also does not improve the clean data-based system or decreases accuracy.
In particular, when all crawled data is used together with the clean data, the MT systems improve their accuracy only 50\% of what they do when trained on our smaller selected data.

 \subsubsection{WMT-18 Shared Task: Machine Translation of News}
To compare with the state-of-the-art, we needed a powerful model, which can approach the results of the best MT systems.
Thus, we used the Transformer ~\cite{NIPS2017_7181}, a more expensive model in terms of computation than the LSTM-based but it is still largely less costly than the top performant systems in the WMT competition.

We trained our MT model with the clean data and the top 6M pairs from ParaCrawl selected with our classifier.
We follow the typical model building pipeline described in \cite{DBLP:journals/corr/abs-1804-00344}.
We use the setting from Marian toolkit~\footnote{\url{https://github.com/marian-nmt/marian-examples/tree/}\\\url{336740065d9c23e53e912a1befff18981d9d27ab/wmt2017-transformer}}.
Table~\ref{main} shows the result.
We note that our model, which uses a relatively much simpler neural network than the state-of-the-art approaches, e.g., RWTH and Microsoft Research (using a Big Transformer), is just 1.6  BLEU score points behind.
This shows that our approach can build more efficient models with less data since the crawled data we used is closer to the target domain.

\begin{table}[t]
\centering
\caption{Comparison of our model with the results reported by WMT-18 using the BLEU score.}\label{main}
\resizebox{\linewidth}{!}{
\begin{tabular}{l c c c c c c}
\toprule
System  & \shortstack{clean\\ pairs} & \shortstack{noisy\\ pairs} & \begin{tabular}[c]{@{}l@{}}monolingual for\\ back-translation\end{tabular} & model & EN-DE & DE-EN \\ %\hline
\midrule
RWTH Aachen  & 6M & 18M & 18M     & Trans.-Big &--& 48.4 \\ %\hline
Microsoft Research  & 6M & 10M & 10M     & Trans.-Big & 48.3 & --\\ %\hline
University of Cambridge & 6M & 15M & 20M     & Trans.-Big & 46.6 & 46.8 \\ %\hline
University of Edinburgh & 6M & 4M & 20M     & Trans.-Base & 44.4 & 43.9 \\ %\hline
JHU MT Systems  & 6M & All & UNK     & RNN & 43.4 & 45.3 \\ %\hline
\shortstack{Universitat Polit{\`e}cnica\\ de Val{\`e}ncia} & 6M & 10M & 20M     & Trans.-Base &--& 45.1 \\ %\hline
\hline
\textbf{Our Model}  & 6M & 6M & 10M     & Trans.-Base & 46.7 & 46.1 \\ %\hline
\bottomrule
\end{tabular}
}
\end{table}

 \subsubsection{Discussion}

Besides automatic evaluation, the WMT-18 Shared Task also conducted  a human evaluation of the participating systems.
Specifically, translations from individual systems were manually validated by assessors, comprised of both researchers and crowd-sourced workers from Mechanical Turk.
The assessment was based on how well a translation replicates the meaning of the reference translation.
The scores from an assessor are first standardized individually, according to their overall mean and standard deviation.
Then, the average standardized scores for translations rated by an assessor for a system are computed.
The overall score, Average $z$, is finally computed as the average of its scores from the assessors.

Table~\ref{humanw18} shows a human evaluation carried out by WMT-2018 organizers.
They consider the systems in Table~\ref{main} and five anonymized commercial translation services, named ONLINE-A, B, C, D and E.
We note that the ranking produced by the manual evaluation is close to the one automatically carried out with BLEU score reported in Table~\ref{main}.
Most critically, the table also shows that almost all online services underperform the top MT participant systems, which are comparable to our approach. 

This is an important comparison as it indirectly shows that the results of our approach are better than those of the services mentioned above.
Additionally, the news domain is not under-represented in MT domains, suggesting that a larger gap between our approach and MT services could be observed when dealing with more specific domains.
In other words, translations from online services may consider moving toward customization, not only for better translations~\cite{dinu-etal-2019-training} but also for better satisfying requests of different groups of {\customer}s.

\section{Related Work}

Previous work has studied methods for selecting effective data for MT.
Some of the approaches include:

    \begin{itemize}
        \item perplexity-based selection: this approach ranks sentences based on the perplexity scores given by a targeted language model~\cite{Gao:2002:TUA:595576.595578,I08-2088,moore-lewis:2010:Short}.
        Only sentences within a certain perplexity threshold are selected.
        \item Language model and translation model combination: this approach ranks sentence-pairs by both the target language model and the translation model trained by general and specific data~\cite{D11-1033,P14-2093}.
        The selection is based on the total cross-entropy difference from both sides.        
    \end{itemize}

The core difference with our proposed approach is that we use
    (i) documents (or at least grouped-sentences) rather than individual sentences~\cite{chen-huang-2016-semi}, and
    (ii) negative examples randomly selected from  a heterogeneous dataset from the web.
    
In contrast with methods aiming at selecting sentences with the same language models, our approach selects documents and thus sentences that belong to the same topics, i.e., approaching the data distribution of specific domains.
In particular, the use of statistics of an entire large document enables a much more robust approach and an accurate selection of data related to the target domain.

Finally, the role of negative examples is also fundamental as patterns present in negative documents are automatically filtered out by the machine learning approach together with the negative sentences.

The business advantage of our approach is clear: given a customer request, we only require their monolingual examples in the target domain, e.g., their websites, documentations, etc.
A classifier for selecting similar training data can be automatically built on their data, as we generate negative examples from the crawled data.
We then apply the classifier to select parallel data from a large repository of parallel data from the Web.
Finally, we train an MT model using the selected data, to obtain a system specialized on the target customer data.
This model, being trained on the target domain data, will generate translations using style and text construction typical from the target domain.
In addition to language customization our  approach also enables the use of smaller models, which have less hardware requirement to fulfill the needs of small or medium enterprises.

\section{Conclusion}
We have proposed our strategy for customizing MT systems' training using data selected from a heterogeneous parallel corpus.
This way, customers can provide their data as examples of the text on which the MT system should provide high accurate translations.
Specifically, we propose a supervised classifier trained on a small sample of monolingual target data.
The classifier makes predictions per batch of sentences to better capture the target domain's patterns and terms.

We show the effectiveness of our method by comparing it with the state-of-the-art on well-known MT benchmarks.
The results demonstrate that we can achieve competitive performance on WMT-18 Shared Tasks, but our approach only requires a small monolingual sample of the target data.
Finally, we believe our proposed method can be applied to customize other IR or Natural Language Processing applications exploiting Web data and IR techniques.

In the future, we are exploring the possibility to apply our method for selecting locale-sensitive training data and thus building locale-specific translation engines.
We will also explore other data dimensions that are orthogonal to the topical categories.
Indeed, we can build a classifier to select particular text styles, ranging from formal (thus building MT systems for translating formal documents), to informal languages, e.g., for more colloquial or less formal text applications, such as blog translation.
We may also be able to target sublanguages and jargons as we can train the MT system with such kind of data, e.g., forums, or non native speaker languages.
We can also build more powerful data selection classifiers that can be learned on customer data in different languages, i.e., neural multilingual topic/style classifiers.

\bibliographystyle{splncs04}
\bibliography{mta}

\end{document}